\documentclass[runningheads]{llncs}

\usepackage[T1]{fontenc}
\usepackage{graphicx}
\usepackage{amsmath,amssymb,amsfonts}
\usepackage{algorithm}
\usepackage{algpseudocode}
\usepackage{cite}
\usepackage{url}
\usepackage{textcomp}
\usepackage{makecell}
\usepackage{adjustbox}

\begin{document}

\title{XSSR: Cross-Domain Self-Supervised Representative Selection for Efficient Annotation in Medical Image Segmentation}

\titlerunning{XSSR: Cross-Domain Representative Selection}

\author{
Byunghyun Ko\inst{1}\textsuperscript{*} \and
Aleksei Anisimov\inst{1}\textsuperscript{*} \and
Kobe Ke\inst{2} \and
Suhas Bharthepude\inst{2} \and
Jeongkyu Lee\inst{2}
}

\authorrunning{B. Ko et al.}

\institute{
Northeastern University, San Jose, CA 95113, USA \\
\email{\{ko.by,anisimov.a\}@northeastern.edu}
\and
Northeastern University, New York, NY 10021, USA \\
\email{\{ke.ko,bharthepude.s,jeo.lee\}@northeastern.edu}
}

\maketitle

\begingroup
\renewcommand{\thefootnote}{}
\footnotetext{$*$ Equal contribution.}
\footnotetext{
Accepted to the Third International Conference on AI in Healthcare (AIiH 2026).
To appear in Springer proceedings.
}
\endgroup

\begin{abstract}
Acquiring labeled medical image data is resource-intensive and a challenge further exacerbated in cross-domain scenarios where source and target datasets differ in imaging equipment, population, or clinical site. This study introduces XSSR (Cross-Domain Self-Supervised Representative Selection), a framework designed to minimize annotation effort in the target domain while maintaining robust segmentation performance. XSSR comprises three stages: first, a Masked Autoencoder (MAE) is trained on unlabeled source data to establish a shared embedding space without requiring target labels; second, a greedy selection algorithm scores unlabeled target samples based on a composite density, novelty, and diversity criterion; and third, a U-Net segmentation model is trained exclusively on the selected subset. The novelty–diversity trade-off parameter, alpha, is automatically calibrated by minimizing embedding-space coverage, eliminating manual tuning. We evaluate XSSR on three public benchmarks: Chest X-ray, RIGA+ retinal fundus imaging, and multi-site Prostate MRI, each under a fixed 5\% annotation budget. XSSR achieves 99.3\% of full-data performance on Chest X-ray using only 22 labeled samples, surpasses random selection by up to 2.5 Dice points on Prostate MRI, and consistently outperforms the CoreSet baseline by 0.4 to 1.2 Dice points across all datasets. Ablation studies indicate that diversity is the most influential scoring component, and per-site analysis shows that performance correlates with scanner similarity to the source domain.

\keywords{Medical image segmentation \and Self-supervised learning \and Masked Autoencoder \and cross-domain adaptation \and Sample selection \and Domain shift \and Data-efficient training \and Annotation efficiency}
\end{abstract}
\section{Introduction}

Medical image segmentation plays an essential role in clinical practice. Modern segmentation tools allow clinicians to automate tasks such as analyzing chest X-rays and MRI scans. Deep learning models, such as U-Net and its variants~\cite{ronneberger2015unet,Cicek2016_3DUNet,oktay2018attentionunetlearninglook}, have achieved high performance when trained and tested on data from the same distribution. However, these deep learning models’ performance often drops in cross-domain scenarios, where differences in imaging devices, patient populations, and other aspects introduce domain shift between training and deployment data.

Acquiring large sets of labeled medical image data remains a major bottleneck for the clinical deployment of deep learning models~\cite{rayed2024survey}. Annotation of medical images is labor- and time-intensive, requiring trained professionals to spend substantial time to annotate images that will be used for training. In cross-domain situations, this challenge becomes even greater, because an entire new dataset needs to be labeled despite the availability of related data. As a result, there is a growing need for models that generalize well across domains while reducing the need for large-scale annotation.

Prior work related to domain adaptation and transfer learning attempts to address the cross-domain issue in medical image segmentation by modifying the image itself or aligning feature distributions~\cite{ganin2016dann,tzeng2017adda,hoffman2018cycada}. While these approaches are effective, they rely on complex, resource-intensive training and assume access to large datasets. In addition, they do not address the issue of annotation efficiency, which is how the model can achieve high segmentation accuracy while minimizing the amount of labeled data required.

Active learning and sample selection methods aim to identify the most informative and representative samples for annotation~\cite{yoo2019learningloss,sener2018active,ash2020badge}. However, most of these approaches are for single-domain environments. As a result, when working with datasets with domain shift, these approaches may ignore images that are both representative of the target domain and sufficiently distinct from the source domain.

To address these challenges, we propose XSSR (Cross-Domain Self-Supervised Representative Selection), a framework that combines self-supervised representation learning with cross-domain representative selection. We first train a Masked Autoencoder (MAE)~\cite{he2022mae} on unlabeled source data to learn the source dataset's feature space and capture its anatomical structure. Self-supervised representation learning methods have been shown to produce strong transferable embeddings for downstream tasks~\cite{chen2020simclr,he2020moco}. The embedding space produced by the MAE is then used to guide the selection of unlabeled target samples through a scoring function that accounts for density, novelty, and diversity. This leads to selecting the most informative target samples for annotation. For improved robustness, we also introduce an automatic calibration mechanism that adaptively balances novelty and diversity without manual tuning. 

We evaluate XSSR on multiple medical imaging benchmarks under different annotation budgets. Our results show that XSSR achieves performance comparable to training with a full dataset while using only a small fraction of labeled images.

The main contributions of this work are summarized as follows:
\begin{itemize}
    \item A cross-domain representative selection framework that reduces annotation cost under domain shift.
    \item Self-supervised MAE training on unlabeled source data to obtain transferable embeddings.
    \item A composite density--novelty--diversity scoring strategy with automatic calibration.
    \item Experimental validation across three medical imaging benchmarks under limited annotation budgets.
\end{itemize}

\section{Related Work}

Recently, deep learning techniques have been widely used for medical image segmentation tasks. Models such as U-Net~\cite{ronneberger2015unet} and its variants~\cite{Cicek2016_3DUNet} have achieved strong segmentation performance on chest X-ray datasets and other medical imaging problems. Attention-based versions further improve results by helping the model focus on important body regions in the image~\cite{oktay2018attentionunetlearninglook}. While these approaches can achieve high Dice similarity scores when trained and tested on the same dataset, their performance often decreases when applied to images from different hospitals or imaging systems. Although current segmentation models perform well on familiar data, they generally do not address problems caused by differences between datasets or the high cost of labeling new data from unseen sources.

To reduce the effects of differences between datasets, domain adaptation and transfer learning methods have been proposed. Some approaches try to make features from the source and target datasets more similar during training~\cite{ganin2016dann,tzeng2017adda}, while other methods aim to reduce visual differences between datasets by learning mappings between them~\cite{hoffman2018cycada}. These techniques can improve performance across different datasets without requiring fully labeled target data, but they often involve complicated training processes and can be hard to train reliably. In addition, most domain adaptation methods assume access to large amounts of unlabeled target data and focus on matching features rather than reducing the number of samples that need to be manually labeled. As a result, they do not directly reduce annotation cost, which remains a major challenge in medical imaging applications.

On the other hand, active learning and sample selection methods aim to reduce labeling effort by choosing only the most useful samples for annotation. Common strategies include selecting samples the model is least confident about~\cite{yoo2019learningloss}, choosing diverse samples~\cite{sener2018active}, or selecting a subset that represents the overall dataset~\cite{ash2020badge}. However, these methods are usually designed for single-dataset settings and do not clearly account for differences between source and target data. When large differences exist between datasets, confidence-based selection may become unreliable, and diversity-based selection alone may fail to choose samples that are truly different from the source data. These limitations motivate the proposed XSSR framework, which combines self-supervised feature learning~\cite{chen2020simclr,he2020moco} with selection based on new and varied samples to address both dataset differences and annotation cost in medical image segmentation.

\section{Methods}

\begin{figure}[t]
\centering
\includegraphics[width=\textwidth]{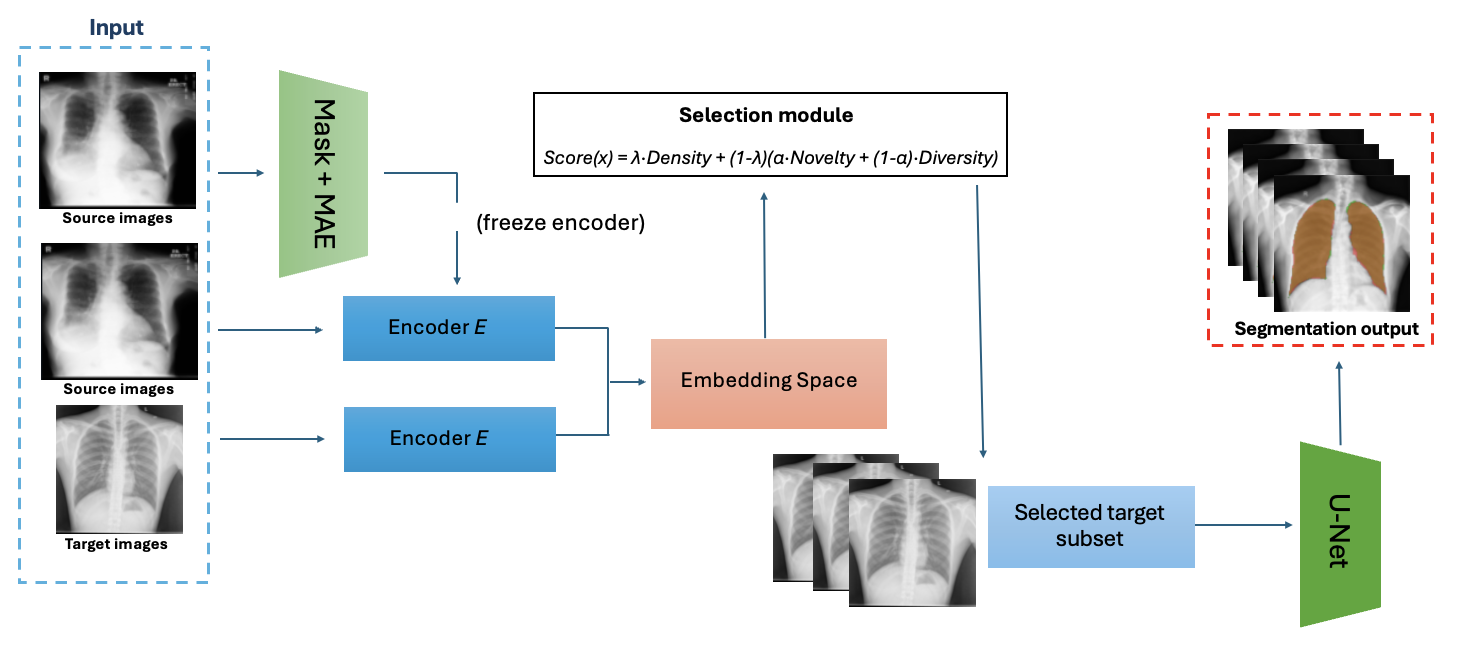}
\caption{Pipeline of the proposed XSSR approach.}
\label{fig:xssr_pipeline}
\end{figure}

We introduce XSSR (Cross-Domain Self-Supervised Representative Selection), a framework designed for data-efficient training under domain shift (see Fig.~\ref{fig:xssr_pipeline}). Given a labeled source dataset $D_s$ and an unlabeled target dataset $D_t$, the goal is to select a subset of target samples that both represent the target distribution and capture differences from the source domain.

The framework consists of three stages:
\begin{itemize}
    \item {\textbf{Stage 1}: Self-supervised representation learning:} learning a feature space using a convolutional masked autoencoder, inspired by MAE~\cite{he2022mae}.
    \item {\textbf{Stage 2}: Greedy sample selection:} selecting informative target samples using an embedding-based scoring function.
    \item {\textbf{Stage 3}: Segmentation model training:} training a segmentation network using only the selected subset.
\end{itemize}

\subsection{Self-Supervised Representation Learning}

A convolutional masked autoencoder, inspired by MAE~\cite{he2022mae}, is trained on the source dataset to learn a feature space that captures anatomical structure without requiring labels. Unlike the original ViT-based MAE, we adopt a lightweight CNN encoder--decoder, which we found more data-efficient for the modest source set sizes typical in medical imaging. Masking is applied directly in pixel space, and the encoder is trained to reconstruct the original image from the masked input.

\subsubsection{Encoder.}
Let $x \in \mathbb{R}^{C \times H \times W}$ denote an input image, where $C$ is the number of channels and $H, W$ are spatial dimensions. The encoder $E(\cdot)$ consists of four stride-2 convolutional blocks that produce a spatial feature map $E_{\text{conv}}(x) \in \mathbb{R}^{d \times H/16 \times W/16}$. We obtain a vector embedding by applying global average pooling (GAP) over the spatial dimensions:
\begin{equation}
z = \operatorname{GAP}(E_{\text{conv}}(x)) \in \mathbb{R}^{d},
\label{eq:encoder}
\end{equation}
where $d$ is the embedding dimension.

\subsubsection{Reconstruction Objective.}
During training, a masked version of the input, denoted $x_{\text{masked}}$, is passed through the autoencoder. The model $f_{\theta}$, with parameters $\theta$, reconstructs the original image using an $L_1$ loss:
\begin{equation}
\mathcal{L}_{\text{MAE}} = \left\| f_{\theta}(x_{\text{masked}}) - x \right\|_1,
\label{eq:mae_loss}
\end{equation}
where $\|\cdot\|_1$ denotes the element-wise $L_1$ norm.

After training, the encoder $E(\cdot)$ is used to extract embeddings for both source and target images:
\begin{equation}
Z_s = \{z_j^s\}_{j=1}^{N_s}, \qquad Z_t = \{z_i^t\}_{i=1}^{N_t},
\label{eq:embedding_sets}
\end{equation}
where $N_s$ and $N_t$ denote the number of source and target samples, respectively.

\subsection{Greedy Sample Selection}

Given the target embedding set $Z_t = \{z_i\}_{i=1}^{N_t}$, where $z_i \in \mathbb{R}^d$ denotes the embedding of target sample $x_i$, the goal is to select a subset $S \subseteq \{1, \dots, N_t\}$ of size $k$.

The selection is performed using a greedy strategy that iteratively adds the highest-scoring sample.

\begin{algorithm}[h]
\caption{Greedy Sample Selection}
\begin{algorithmic}[1]
\State \textbf{Input:} target embeddings $Z_t$, source embeddings $Z_s$, budget $k$
\State \textbf{Initialize:} $S \leftarrow \emptyset$
\For{$t = 1$ to $k$}
    \State Compute $\text{Score}(x_i)$ for all $i \notin S$
    \State $i^* \leftarrow \arg\max_{i \notin S} \text{Score}(x_i)$
    \State $S \leftarrow S \cup \{i^*\}$
\EndFor
\State \textbf{Return:} selected index set $S$
\end{algorithmic}
\end{algorithm}

Here, $S$ denotes the set of selected sample indices, and $\text{Score}(x_i)$ is defined in Equation 4. The density and novelty terms are precomputed for all samples, while the diversity term depends on $S$ and is updated at each iteration.

\subsubsection{Selection Score.}
Each target sample $x_i$ is associated with an embedding $z_i \in Z_t$. The overall selection score is defined as:
\begin{equation}
\begin{aligned}
\textit{Score}(x_i) = {} & \lambda \cdot \textit{Density}(x_i) \\
& + (1 - \lambda)\left[\alpha \cdot \textit{Novelty}(x_i) + (1 - \alpha)\cdot \textit{Diversity}(x_i)\right],
\end{aligned}
\label{eq:score}
\end{equation}
where $\lambda \in [0,1]$ controls the importance of density and $\alpha \in [0,1]$ balances novelty and diversity. All score components are normalized using z-score normalization before combination.

\subsubsection{Density.}
Density measures how representative a sample is within the target dataset. For each target embedding $z_i$, let $\mathcal{N}_k(i)$ denote the set of its $k$ nearest neighbors in $Z_t$ excluding itself. The average neighbor distance is:
\begin{equation}
d_i = \frac{1}{k} \sum_{j \in \mathcal{N}_k(i)} \left\| z_i - z_j \right\|_2.
\label{eq:density_dist}
\end{equation}
Density is defined as the negative mean distance:
\begin{equation}
\textit{Density}(x_i) = -d_i.
\label{eq:density}
\end{equation}
Samples located in dense regions, corresponding to small $d_i$, receive higher scores.

\subsubsection{Diversity.}
Diversity encourages coverage of different regions of the target embedding space. Let $S$ denote the set of selected samples. Diversity is defined as the distance from $z_i$ to the nearest selected embedding:
\begin{equation}
\textit{Diversity}(x_i) = \min_{x_j \in S} \left\| z_i - z_j \right\|_2.
\label{eq:diversity}
\end{equation}
Samples far from the current selection receive higher diversity scores.

\subsubsection{Novelty.}
Novelty measures how different a target sample is from the source domain. Let $Z_s$ denote the set of source embeddings. Novelty is defined as:
\begin{equation}
\textit{Novelty}(x_i) = \min_{z_j^s \in Z_s} \left\| z_i - z_j^s \right\|_2,
\label{eq:novelty}
\end{equation}
where $z_j^s$ is a source embedding. Higher values indicate greater domain shift relative to the source dataset.

\subsubsection{Automatic Calibration of $\alpha$.}
The parameter $\alpha$ controls the trade-off between novelty and diversity. Instead of fixing it manually, we determine $\alpha$ automatically. For each candidate value of $\alpha$, we (1) run the greedy selection algorithm to obtain a subset $S$, and (2) measure coverage of the target dataset as:
\begin{equation}
\textit{Coverage} =
\frac{1}{N_t} \sum_{i=1}^{N_t} \min_{x_j \in S} \left\| z_i - z_j \right\|_2,
\label{eq:coverage}
\end{equation}
where $N_t$ is the number of target samples. The value of $\alpha$ that minimizes coverage is selected.

\subsection{Segmentation Model Training}

After selection, a segmentation model $f_{\theta}$, which is implemented as a U-Net, is trained using only the selected subset $S$. Given an input image $x$ and ground-truth segmentation mask $y$, the model is optimized using binary cross-entropy loss:
\begin{equation}
\mathcal{L}_{\text{seg}} = \textit{BCE}(f_{\theta}(x), y),
\label{eq:seg}
\end{equation}
where $\textit{BCE}(\cdot)$ denotes the binary cross-entropy function. Training is performed exclusively on the selected subset, significantly reducing annotation cost while maintaining performance.

\section{Experimental Design}

\subsection{Baselines}

To evaluate the effectiveness of the proposed selection strategy, we compare against: (1) full-data training, using all target samples; (2) random selection, where $k$ samples are randomly chosen; and (3) CoreSet selection, a diversity-based method using greedy farthest-point sampling.

\subsection{Dataset}

Three publicly available medical image segmentation datasets are used: Chest X-ray, retinal fundus images (RIGA+), and prostate MRI.

The \textbf{Chest X-ray} dataset uses the Montgomery County chest X-ray set~\cite{jaeger2014montgomery} as the source domain and the Shenzhen Hospital chest X-ray set~\cite{jaeger2014montgomery} as the target domain. The full training set contains 453 labeled images. The \textbf{RIGA+} retinal fundus imaging dataset~\cite{al2015riga} is used for optic disc segmentation. BinRushed and Magrabia are used as the source domains, while MESSIDOR images are used as the target domain, with 362 labeled training images. The \textbf{Prostate MRI} benchmark follows the commonly used six-site multi-domain prostate MRI benchmark introduced in prior cross-domain segmentation work~\cite{liu2020msnet,samlbenchmark}. The benchmark aggregates institutional cohorts originating from the NCI-ISBI 2013 challenge (RUNMC and BMC), the I2CVB cohort defined in the SAML benchmark, and PROMISE12-derived cohorts (UCL, HK, and BIDMC)~\cite{litjens2014prostate,bloch2015nciisbi}. RUNMC is used as the source domain, while the remaining five sites are treated as target domains. Following prior work, all MRI volumes are preprocessed into 2D slices for training and evaluation. Annotation budgets are computed at the 2D slice level after preprocessing. The full training set contains 1159 labeled slices across all sites. In all experiments, only 5\% of the images in the target domain are annotated for training.

\subsection{Evaluation Metrics}

The evaluation metric for all experiments is the Dice similarity coefficient (DSC), which ranges from 0 to 1 with scores closer to 1 indicating better agreement with ground truth. Model outputs are binarized using a threshold of 0.5.

\subsection{Comparison}

We compare XSSR against CoreSet~\cite{sener2018active} and random selection at a 5\% annotation budget, $k=22$ for Chest X-ray, $k=18$ for RIGA+, $k=57$ for Prostate MRI. All methods train a U-Net with identical hyperparameters: 100 epochs, batch size of 8, learning rate of $1\times10^{-3}$, AdamW optimizer, and input resolution of $128\times128$ pixels.

\section{Experimental Results}

\subsection{Method Comparison}

Table~\ref{tab:baseline} summarizes segmentation performance when training on all available target-domain labels, establishing the upper-bound reference for each dataset. Chest X-ray involves a grayscale domain shift between Montgomery County (USA) and Shenzhen (China) hospitals. Fundus (RIGA+) involves RGB retinal imaging across different clinical sites for optic disc segmentation. Prostate MRI presents the most challenging scenario, with a single Siemens 3T source scanner and five heterogeneous target sites spanning three manufacturers and two field strengths.

\begin{table}[htbp]
\caption{Upper-Bound Performance (100\% Annotated Target Data).}
\label{tab:baseline}
\centering
\begin{tabular}{|c|c|c|c|c|}
\hline
\textbf{Dataset} & \textbf{Modality} & \textbf{Source $\rightarrow$ Target} & \textbf{N} & \textbf{Dice} \\
\hline
Chest X-ray & Grayscale & Montgomery $\rightarrow$ Shenzhen & 453 & 0.952 \\
\hline
Fundus (RIGA+) & RGB & BinRushed $\rightarrow$ MESSIDOR & 362 & 0.974 \\
\hline
Prostate MRI & Grayscale & RUNMC $\rightarrow$ 5 sites & 1159 & 0.918 \\
\hline
\end{tabular}
\end{table}

Table~\ref{tab:selection} compares XSSR against CoreSet~\cite{sener2018active} and uniform random selection. All methods operate at 5\% annotated data, meaning each method selects $k$ samples from the unlabeled target pool for annotation, where $k = \lfloor 0.05 \times N \rfloor$ and $N$ is the total number of available target-domain images. The remaining 95\% of the target data is never annotated. XSSR outperforms both alternatives on all three datasets. On Chest X-ray, selecting only $k$=22 out of 453 images achieves 0.945 Dice, retaining 99.3\% of the upper bound while reducing annotation effort by 95\%. On Fundus, $k$=18 selected images reach 0.940 Dice, improving over random selection by 1.3 Dice points. The largest margin appears on Prostate MRI, where XSSR's $k$=57 selected slices yield 0.760 Dice compared to 0.735 for random, a gain of 2.5 Dice points. This wider gap reflects the greater difficulty of covering five heterogeneous scanner domains with limited samples.

\begin{table}[htbp]
\caption{Method Comparison at 5\% Annotated Data. The ``vs Rand'' column reports the absolute Dice improvement of XSSR over random selection.}
\label{tab:selection}
\centering
\begin{tabular}{|c|c|c|c|c|c|c|}
\hline
\textbf{Dataset} & \textbf{k} & \textbf{XSSR} & \textbf{CoreSet} & \textbf{Random} & \textbf{vs Rand ($\Delta$ Dice)} & \textbf{\% Base} \\
\hline
Chest X-ray & 22 & \textbf{0.945} & 0.941 & 0.938 & +0.007 & 99.3\% \\
\hline
Fundus & 18 & \textbf{0.940} & 0.935 & 0.927 & +0.013 & 96.5\% \\
\hline
Prostate MRI & 57 & \textbf{0.760} & 0.748 & 0.735 & +0.025 & 82.8\% \\
\hline
\end{tabular}
\end{table}

Compared to CoreSet, XSSR achieves 0.4 to 1.2 Dice points higher across all datasets. CoreSet performs greedy farthest-point sampling in embedding space, which maximises geometric spread but tends to select outliers. XSSR mitigates this through its density component, which steers selection toward representative regions of the target distribution while still maintaining coverage.

\begin{figure}[htbp]
\centerline{\includegraphics[width=0.85\columnwidth]{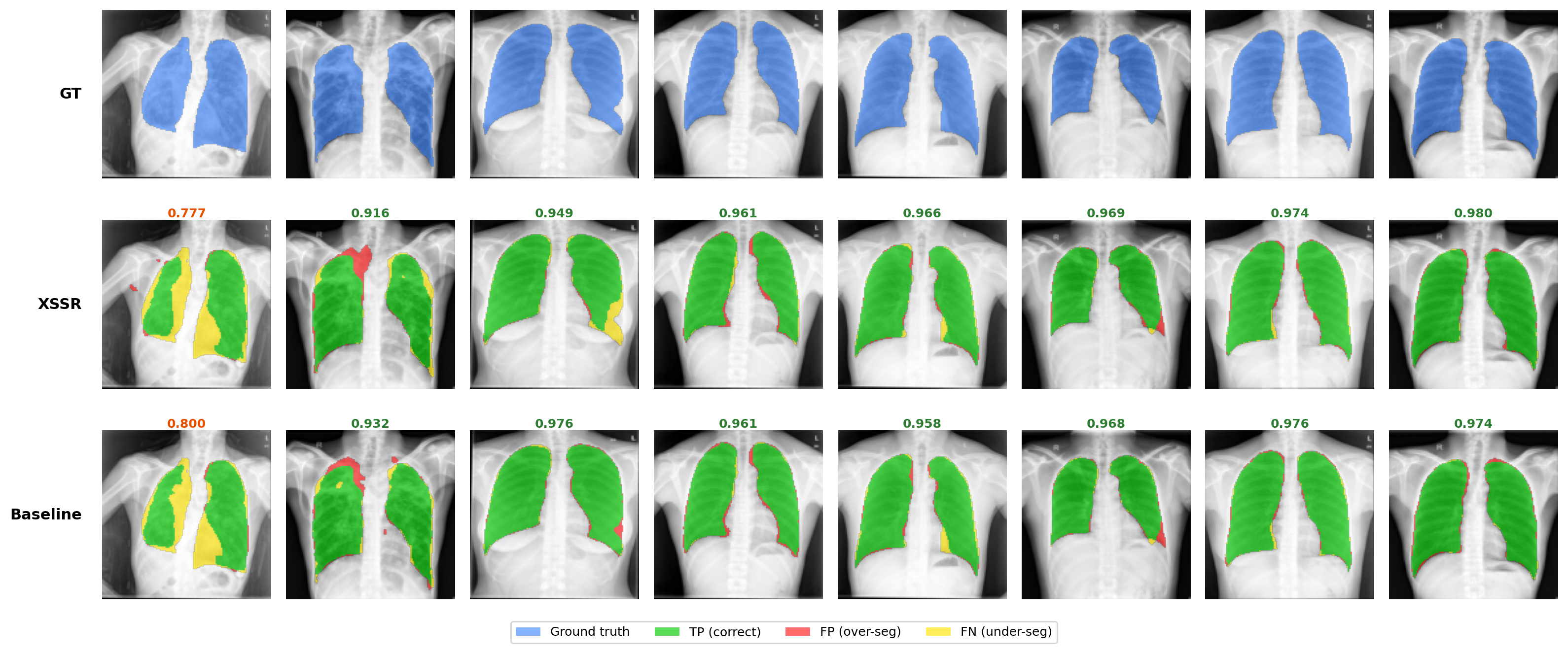}}
\caption{Chest X-ray lung segmentation. Top row: ground truth (blue). Middle row: XSSR (22 samples, 5\%). Bottom row: baseline (100\%). Predictions use TP/FP/FN color coding: green = correct, red = over-segmentation, yellow = under-segmentation. Samples sorted by XSSR Dice from worst (left) to best (right).}
\label{fig:overlay_xray}
\end{figure}

\begin{figure}[htbp]
\centerline{\includegraphics[width=0.85\columnwidth]{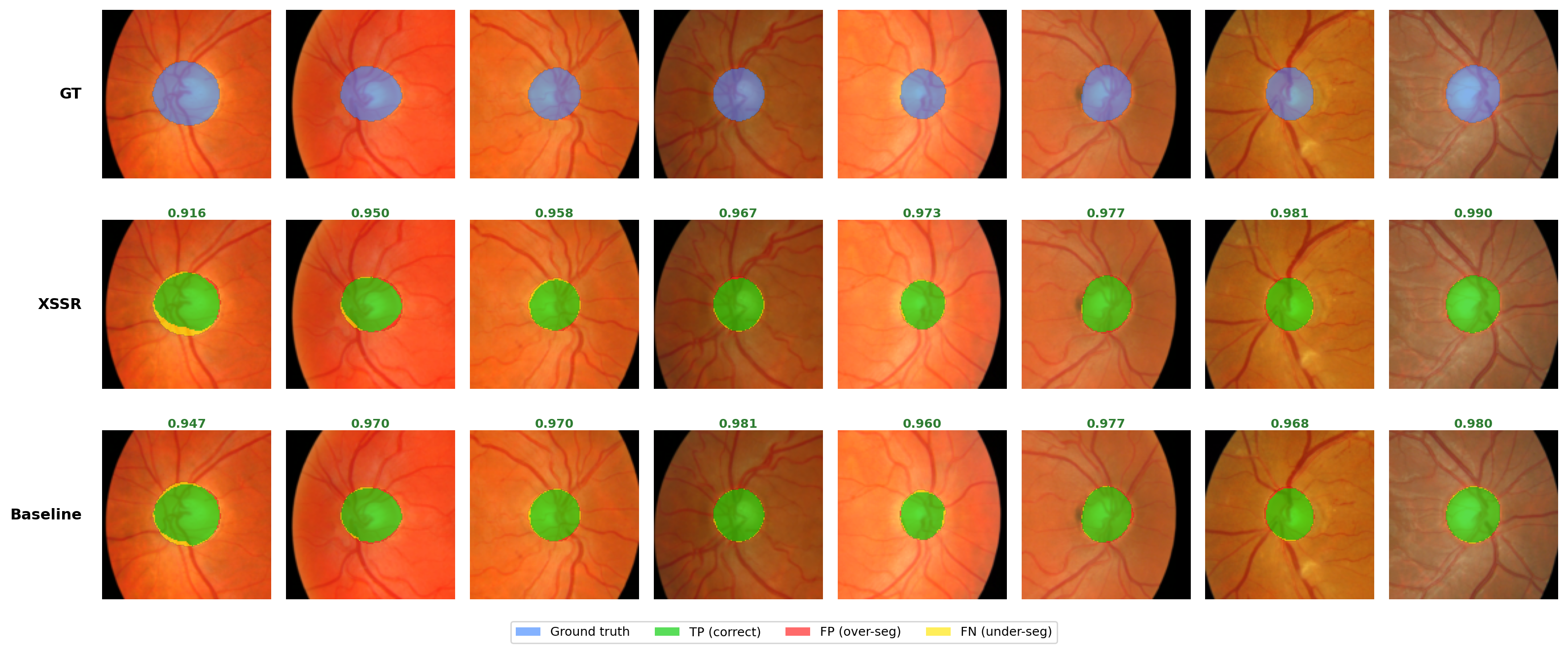}}
\caption{Fundus optic disc segmentation. Rows and color coding as in Fig.~\ref{fig:overlay_xray}; XSSR uses 18 samples (5\%).}
\label{fig:overlay_riga} 
\end{figure}

\begin{figure}[htbp]
\centerline{\includegraphics[width=0.85\columnwidth]{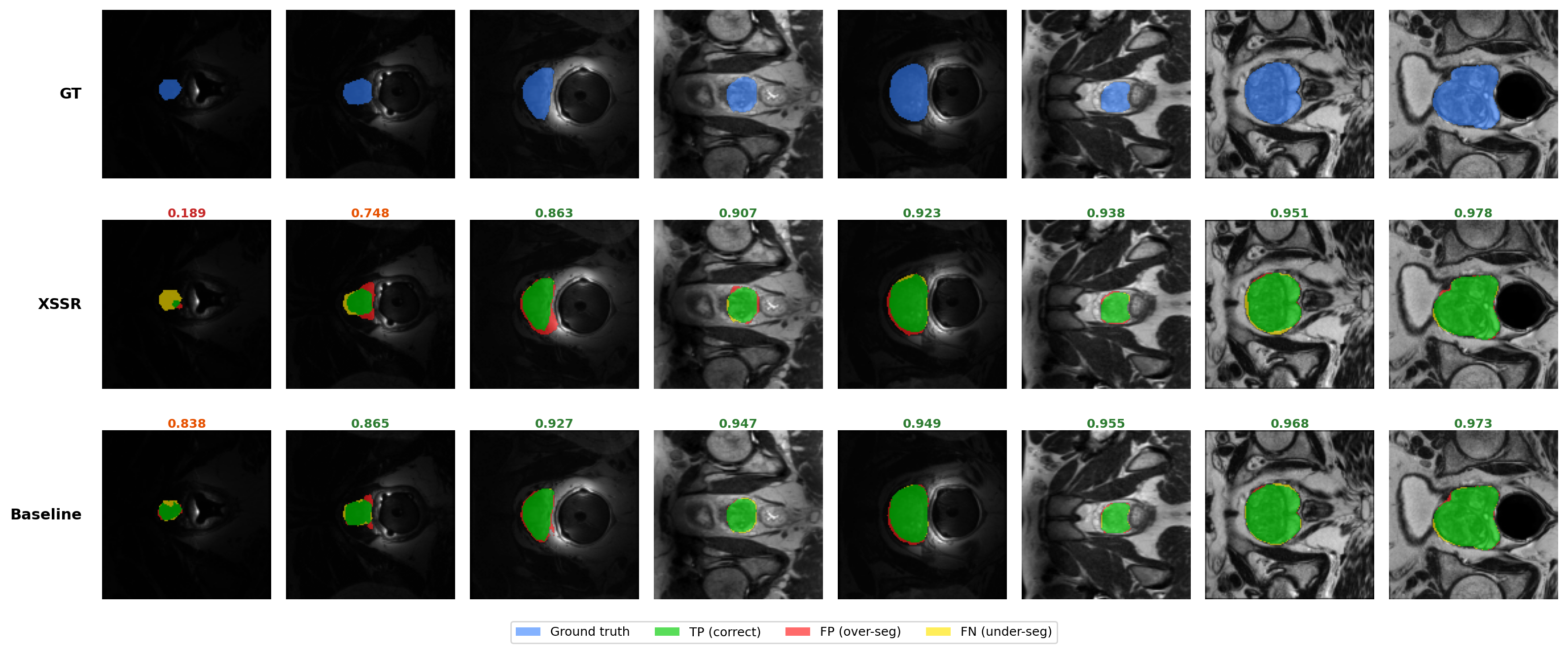}}
\caption{Prostate MRI segmentation. Rows and color coding as in Fig.~\ref{fig:overlay_xray}; XSSR uses 57 slices (5\%).}
\label{fig:overlay_prostate}
\end{figure}

Figures~\ref{fig:overlay_xray} to~\ref{fig:overlay_prostate} show qualitative segmentation results for each dataset. Despite training on only 5\% of the available target annotations, XSSR produces predictions that closely match the ground truth across all three tasks.

\subsection{Robustness Across Domain Shifts}

To evaluate how XSSR handles varying degrees of domain shift, Table~\ref{tab:persite} reports per-site results on Prostate MRI at 5\% annotated data ($k$=57). The source domain is RUNMC (Siemens 3T with surface coil), and each target site differs in scanner manufacturer, field strength, or coil type.

\begin{table}[htbp]
\caption{Per-Site Results on the Five Target Sites of the Six-Site Prostate MRI Benchmark ($k$=57).}
\label{tab:persite}
\centering
\begin{tabular}{|c|c|c|c|c|c|}
\hline
\textbf{Site} & \textbf{Scanner} & \textbf{XSSR} & \textbf{Base} & \textbf{\% Base} & \textbf{Shift} \\
\hline
I2CVB & Siemens 3T & 0.904 & 0.952 & 95.0\% & Small \\
\hline
BMC & Philips 1.5T & 0.882 & 0.909 & 97.0\% & Medium \\
\hline
UCL & Siemens 1.5T/3T & 0.855 & 0.872 & 98.1\% & Medium \\
\hline
HK & Siemens 1.5T & 0.832 & 0.897 & 92.8\% & Small--Med \\
\hline
BIDMC & GE 3T & 0.774 & 0.916 & 84.5\% & Large \\
\hline
\end{tabular}
\end{table}

XSSR retains 92.8--98.1\% of upper-bound performance on Siemens and Philips sites, but only 84.5\% on BIDMC (GE 3T with endorectal coil)---the most distinct setup relative to the Siemens source. This shows XSSR handles moderate domain shifts well, but extreme hardware differences remain challenging at very low annotation levels.

\subsection{Ablation Study}

Ablation experiments use a separate run batch from Table~\ref{tab:selection}; small cross-table differences reflect typical seed-to-seed variance and do not affect within-table comparisons.

\subsubsection{Component Analysis.}
Table~\ref{tab:ablation} isolates the contribution of each scoring component by removing one at a time, with all experiments conducted at 5\% annotated data.

\begin{table}[htbp]
\caption{Component Ablation (Dice at 5\% Annotated Data).}
\label{tab:ablation}
\centering
\begin{tabular}{|c|c|c|c|}
\hline
\textbf{Setting} & \textbf{Chest X-ray} & \textbf{Fundus} & \textbf{Prostate MRI} \\
\hline
\textbf{Full XSSR} & \textbf{0.943} & \textbf{0.945} & \textbf{0.732} \\
\hline
without diversity ($\alpha$=1.0) & 0.912 & 0.934 & 0.350 \\
\hline
without density ($\lambda$=0) & 0.938 & 0.917 & 0.716 \\
\hline
without novelty ($\alpha$=0) & 0.940 & 0.905 & 0.743 \\
\hline
\hline
Upper bound (100\% data) & 0.952 & 0.974 & 0.919 \\
\hline
\end{tabular}
\end{table}

The relative importance of each component varies with the nature of the domain shift. On Prostate MRI, diversity is critical: removing it collapses Dice from 0.732 to 0.350, because without explicit spatial coverage the selection concentrates on a single scanner site and fails to generalise across the remaining four. On Fundus, novelty contributes the most---removing it causes the largest single-component drop from 0.945 to 0.905---because the BinRushed-to-MESSIDOR shift involves systematic appearance differences where selecting source-dissimilar samples is essential. On Chest X-ray, all three components contribute, with diversity removal again causing the largest degradation from 0.943 to 0.912.

Density provides a consistent but smaller benefit across all datasets ($+$0.005 to $+$0.028 Dice), confirming that preferring samples from dense target regions prevents the selection of uninformative outliers. Notably, on Prostate MRI the ``without novelty'' variant slightly outperforms full XSSR (0.743 vs.\ 0.732), consistent with the auto-$\alpha$ mechanism already assigning near-zero novelty weight for this dataset. The small residual gap falls within typical run-to-run variance and confirms that the auto-calibration correctly identifies novelty as unhelpful under multi-site shifts.

These dataset-dependent patterns are precisely what motivates the auto-$\alpha$ calibration mechanism: rather than relying on a fixed novelty--diversity trade-off, XSSR adapts the balance to the characteristics of each domain shift.

\subsubsection{Annotation Efficiency.}
Table~\ref{tab:budget} examines how the amount of annotated target data affects the percentage of upper-bound performance retained.

\begin{table}[htbp]
\caption{Percentage of Upper-Bound Retained vs.\ Amount of Annotated Data. Bold indicates the better method per dataset at each level.}
\label{tab:budget}
\centering
\begin{tabular}{|c|cc|cc|cc|}
\hline
& \multicolumn{2}{c|}{\textbf{Chest X-ray}} & \multicolumn{2}{c|}{\textbf{Fundus}} & \multicolumn{2}{c|}{\textbf{Prostate MRI}} \\
\textbf{Annotated} & XSSR & Rand & XSSR & Rand & XSSR & Rand \\
\hline
5\% & \textbf{99.2} & 98.9 & \textbf{95.7} & 94.0 & \textbf{80.0} & 72.6 \\
\hline
10\% & \textbf{99.6} & 99.5 & 96.7 & \textbf{97.8} & 87.8 & \textbf{89.9} \\
\hline
20\% & \textbf{99.8} & 99.8 & 99.1 & 99.1 & \textbf{93.7} & 93.5 \\
\hline
50\% & 99.8 & \textbf{99.9} & 99.7 & \textbf{99.8} & 97.7 & \textbf{98.2} \\
\hline
\end{tabular}
\end{table}

XSSR's advantage over random selection is most pronounced at 5\% annotated data, where intelligent selection matters most: the gap is 0.3 points on Chest X-ray, 1.7 on Fundus, and 7.4 on Prostate MRI. As the annotation pool grows, random sampling gradually covers the target distribution and the gap narrows. On Prostate MRI, random selection slightly exceeds XSSR above 10\% annotated data. This occurs because uniform random sampling naturally draws from all five scanner sites in proportion to their representation in the pool, providing balanced multi-site coverage that XSSR's embedding-driven selection does not explicitly enforce. This suggests that incorporating site-awareness into the selection criterion could further improve XSSR's performance on multi-site datasets at higher annotation levels.

\section{Conclusion}

We introduced XSSR, a cross-domain framework for data-efficient medical image segmentation that combines MAE-based feature learning with a density-novelty-diversity scoring strategy and automatic calibration. Across Chest X-ray, RIGA+, and Prostate MRI under a 5\% annotation budget, XSSR outperforms random sampling and CoreSet, retains up to 99.3\% of full-data performance with 22 labeled samples on Chest X-ray, and beats random selection by up to 2.5 Dice points on Prostate MRI. Ablations confirm diversity as the most influential component, and per-site analysis shows that performance degradation correlates with scanner dissimilarity to the source. Future work includes external validation on additional modalities and embedding refinements for large cross-manufacturer shifts.

\end{document}